\documentclass[sigconf]{acmart}
\acmSubmissionID{578}

\AtBeginDocument{%
  \providecommand\BibTeX{{%
    \normalfont B\kern-0.5em{\scshape i\kern-0.25em b}\kern-0.8em\TeX}}}

\usepackage{environ}
\ExplSyntaxOn
\seq_new:N \g_appendices_seq% define a sequence for holding the appendices
\NewEnviron{Appendix}{\seq_gput_right:No \g_appendices_seq \BODY}
\newcommand\AddAppendices{% regurgitate the appendices
	\appendix% turn all subsequent sections into appendices
	\seq_map_inline:Nn \g_appendices_seq {##1}
}
\ExplSyntaxOff

\usepackage{etoolbox}
\AtEndDocument{\AddAppendices}

\setcopyright{acmcopyright}
\copyrightyear{2021}
\acmYear{2021}
\setcopyright{acmcopyright}\acmConference[MM '21]{Proceedings of the 29th ACM International Conference on Multimedia}{October 20--24, 2021}{Virtual Event, China}
\acmBooktitle{Proceedings of the 29th ACM International Conference on Multimedia (MM '21), October 20--24, 2021, Virtual Event, China}
\acmPrice{15.00}
\acmDOI{10.1145/3474085.3475275}
\acmISBN{978-1-4503-8651-7/21/10}

\usepackage{graphicx}
\usepackage{subfigure}
\usepackage{xcolor}
\usepackage{enumitem}

\newif\ifisfinal
\isfinaltrue

\usepackage{multirow}

\newcommand\NetName{UniCon}

\begin{document}
\fancyhead{}

\title{\NetName{}: Unified Context Network for Robust Active Speaker Detection}

\author{Yuanhang Zhang}
\ifisfinal
\authornote{Both authors contributed equally to this research.}
\fi
\affiliation{%
	\institution{Key Laboratory of Intelligent Information Processing, Institute of Computing Technology, Chinese Academy of Sciences (CAS)}
	\city{Beijing}
	\postcode{100190}
	\country{China}
}
\email{zhangyuanhang15@mails.ucas.ac.cn}
\ifisfinal
\additionalaffiliation{%
	\institution{School of Computer Science and Technology, University of Chinese Academy of Sciences}
	\city{Beijing}
	\postcode{100049}
	\country{China}
}
\fi

\author{Susan Liang}
\ifisfinal
\authornotemark[1]
\authornotemark[2]
\fi
\affiliation{%
	\institution{Key Laboratory of Intelligent Information Processing, Institute of Computing Technology, CAS}
	\city{Beijing}
	\postcode{100190}
	\country{China}
}
\email{liangsusan18@mails.ucas.ac.cn}

\author{Shuang Yang}
\ifisfinal
\authornotemark[2]
\fi
\affiliation{%
	\institution{Key Laboratory of Intelligent Information Processing, Institute of Computing Technology, CAS}
	\city{Beijing}
	\postcode{100190}
	\country{China}
}
\email{shuang.yang@ict.ac.cn}

\author{Xiao Liu}
\affiliation{%
	\institution{Tomorrow Advancing Life}
	\city{Beijing}
	\country{China}
}
\email{ender.liux@gmail.com}

\author{Zhongqin Wu}
\affiliation{%
	\institution{Tomorrow Advancing Life}
	\city{Beijing}
	\country{China}
}
\email{30388514@qq.com}

\author{Shiguang Shan}
\ifisfinal
\authornotemark[2]
\fi
\affiliation{%
	\institution{Key Laboratory of Intelligent Information Processing, Institute of Computing Technology, CAS}
	\city{Beijing}
	\postcode{100190}
	\country{China}
}
\email{sgshan@ict.ac.cn}

\author{Xilin Chen}
\ifisfinal
\authornotemark[2]
\fi
\affiliation{%
	\institution{Key Laboratory of Intelligent Information Processing, Institute of Computing Technology, CAS}
	\city{Beijing}
	\postcode{100190}
	\country{China}
}
\email{xlchen@ict.ac.cn}

\renewcommand{\shortauthors}{Zhang and Liang, et al.}

\begin{abstract}
	We introduce a new efficient framework, the Unified Context Network (\NetName{}), for robust active speaker detection (ASD).
	Traditional methods for ASD usually operate on each candidate's pre-cropped face track separately and do not sufficiently consider the relationships among the candidates. This potentially limits performance, especially in challenging scenarios with low-resolution faces, multiple candidates, etc. 
	Our solution is a novel, unified framework that focuses on jointly modeling multiple types of \textit{contextual} information:
	spatial context to indicate the position and scale of each candidate's face, relational context to capture the visual relationships among the candidates and contrast audio-visual affinities with each other, and temporal context to aggregate long-term information and smooth out local uncertainties. 
	Based on such information, our model optimizes all candidates in a unified process for robust and reliable ASD.
	A thorough ablation study is performed on several challenging ASD benchmarks under different settings. In particular, our method outperforms the state-of-the-art by a large margin of about $15$\% mean Average Precision (mAP) absolute on two challenging subsets: one with three candidate speakers, and the other with faces smaller than $64$ pixels. Together, our \NetName{} achieves $92.0$\% mAP on the AVA-ActiveSpeaker validation set, surpassing $90$\% for the first time on this challenging dataset at the time of submission. Project website: \url{https://unicon-asd.github.io/}.
\end{abstract}

\begin{CCSXML}



\maketitle

\section{Introduction}
\begin{figure*}
	\centering
	\includegraphics[width=\textwidth]{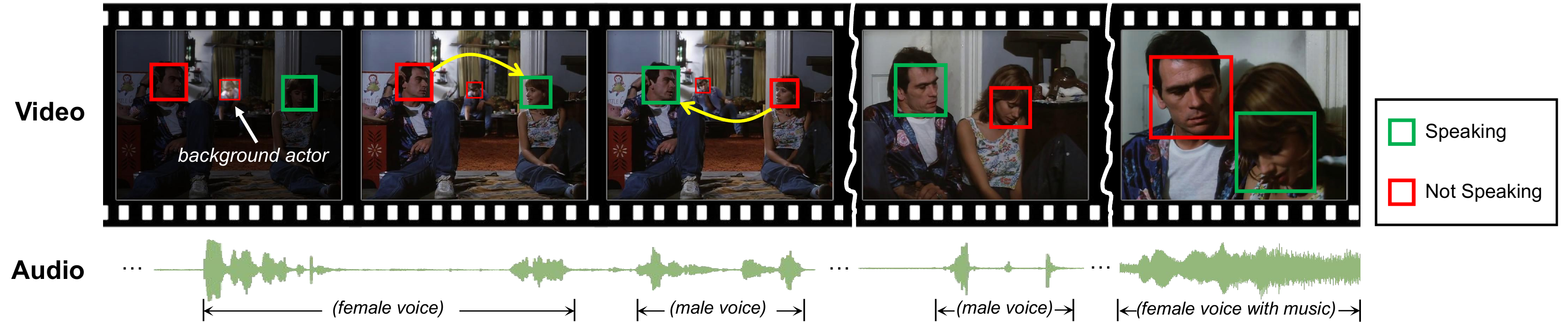}
	\vskip-2ex
	\caption{\textit{What marks an active speaker out from others?} Admittedly, face motion and its synchrony with the audio are the most obvious clues; however, as shown in the figure above, the underlying signals can be highly ambiguous, especially in hard scenarios with poorly lit, low-resolution faces and noisy acoustics, etc. We observe that three additional types of \textit{contextual} clues can help identify the active speaker: (a) higher correspondence between his/her face track and the audio relative to other candidates, (b) a form of contrast with the remaining candidates in terms of visual saliency, the amount of attention he/she receives from others, etc. and (c) temporal continuity. In this paper, we propose a novel Unified Context Network (\NetName{}) which models such context information to jointly optimize all candidates for robust active speaker detection.}
	\label{fig:teaser}
	\vskip-2ex
\end{figure*}

% 1. Frame the research problem
% establish the context, background and/or importance of the topic
Active Speaker Detection (ASD) refers to the task of identifying when each visible person is speaking in a video, typically through careful joint analysis of face motion and voices. It has a wide range of modern practical applications, such as video re-targeting~\cite{DBLP:conf/icassp/CutlerMJZKWK20}, speech diarization~\cite{DBLP:conf/interspeech/ChungHNAZ20}, automatic video indexing, human-robot interaction~\cite{7029977}, speech enhancement, and the automatic curation of large-scale audio-visual speech datasets~\cite{DBLP:journals/tog/EphratMLDWHFR18,DBLP:conf/interspeech/ChungNZ18,DBLP:conf/icassp/FanKLLCCZZCW20}. 

% 2. Mention previous studies on this topic
% giving a brief review of the relevant academic literature
Although research in this direction dates back almost two decades \cite{DBLP:conf/icmcs/CutlerD00,DBLP:conf/nips/SlaneyC00}, most early work only experiment with simple, short sequences depicting frontal faces, which is a major simplification of practical settings. Recent developments in deep learning for audio-visual tasks have led to renewed interest in the ASD problem~\cite{DBLP:conf/accv/ChungZ16a,DBLP:conf/icassp/HooverCPSS18}. In 2020, Roth \textit{et al.}~\cite{DBLP:conf/icassp/RothCKMGKRSSXP20} introduced AVA-ActiveSpeaker, the first large-scale, challenging ASD benchmark extracted from diverse YouTube movies, which significantly boosted subsequent research. Several effective methods have been proposed, building upon 2D and 3D convolutional neural networks (CNNs) and recurrent neural networks (RNNs)~\cite{DBLP:conf/icassp/RothCKMGKRSSXP20,DBLP:journals/corr/abs-1906-10555,zhangmulti2019,DBLP:conf/cvpr/AlcazarCMPLAG20}. 
% 3. Why was this study performed? (The rationale)
% identifying a problem, controversy or a knowledge gap in the field of study
Among them, most closely related to our work is the Active Speaker Context (ASC) model proposed in 2020~\cite{DBLP:conf/cvpr/AlcazarCMPLAG20}, which combines cropped face tracks from all candidates with the audio using a self-attention module~\cite{DBLP:conf/nips/VaswaniSPUJGKP17} to model relationships among the candidates. However, relationships learned in this manner are weak, and lack clear interpretation. In addition, like most existing methods, the ASC model also focuses on optimizing each candidate separately. As a consequence, in complex scenarios with multiple candidates, low-resolution faces, occlusion, noisy acoustics etc., current ASD models often lack robustness and cannot yield satisfactory results. 

% 4. What are the aims and objectives of this study?
% stating the aim(s) of the research 
In this paper, we present a fundamentally different approach where we jointly model multiple sources of \textit{contextual} information and simultaneously optimize all candidates in the scene in a unified process for robust ASD. 
% 5. What was the research hypothesis?
% and the research questions or hypotheses
As shown in Fig. \ref{fig:teaser}, we observe that when face motion and audio signals are ambiguous, three other types of contextual clues often help mark the true active speaker out from others: 
(a) \textit{higher} correspondence between his/her face track and the audio relative to others, e.g. hearing a male's voice helps eliminate female candidates; 
(b) a certain form of contrast with the remaining candidates in terms of his/her visual saliency, the amount of attention he/she received from others, etc., for instance, one person speaks while the rest look at him/her; 
and (c) temporal continuity, \textit{i.e.} by observing each candidate's behavior in a temporal neighborhood, we can smooth out erroneous predictions that arise from instantaneous frame-level decisions.

% providing a synopsis of the research design and method(s)
In light of these observations, we propose a novel framework that seeks to resolve local ambiguities and achieve robust ASD by jointly modeling different kinds of contextual evidence. 
Specifically, we introduce face position and scale information as global \textit{spatial context} to implicitly reflect the visual saliency and the visual focus of attention (VFoA) of each candidate in the scene.
Building upon this information, we then construct a powerful and efficient multi-speaker \textit{relational context} where each candidate is contrasted with others from both visual and audio-visual perspectives. 
Finally, we integrate \textit{temporal context} into the previous relational context to aggregate long-term information and remove noises in frame-level predictions. 
By incorporating spatial, relational, and temporal context into a unified framework, our method can jointly optimize all candidates in the scene end-to-end. In the experiments, we show that our method significantly improves ASD performance, especially in challenging scenarios.

% provide an overview of the coverage and/or structure of the writing
In summary, the main contributions of our work are:
\begin{itemize}[topsep=2pt]
	\item We propose a well-motivated, compact framework, the Unified Context Network (\NetName{}) for robust ASD, which unifies spatial, relational, and temporal context to jointly optimize all candidates in the scene.
	\item We highlight the use of face position and scale in constructing each candidate's global spatial context, and explore the modeling of the relationships among the candidates.
	\item We conduct thorough ablation studies that prove the effectiveness of our method, particularly under challenging multi-speaker and low-resolution settings. Moreover, we outperform the current state-of-the-art by a large margin on challenging ASD benchmarks.
\end{itemize}
\vspace{-2ex}

\section{Related Work}
\paragraph{Active speaker detection (ASD)} Current approaches for ASD can be summarized into two paradigms: unsupervised and supervised. \textit{Unsupervised} methods usually operate under one of two assumptions: (a) the active speaker's face is the one with the highest correspondence between the audio and its visual information, or (b) the active speaker's face is the one that co-occurs most frequently with the person's idiosyncratic voice. Under assumption (a), one typical practice is to learn an audio-visual two-stream network using a contrastive loss between the two modalities~\cite{DBLP:conf/accv/ChungZ16a}. During training, pairs of temporally aligned audio and visual signals are pulled close, while misaligned ones are pushed apart. During inference, the active speaker is naturally the one with lowest distance between audio and visual features. Recently, the method has been improved by better training objective designs, such as multi-way classification~\cite{DBLP:conf/icassp/ChungCK19} and multinomial loss~\cite{DBLP:conf/icassp/DingXZCW20}. Methods based on assumption (b) require strong face and voice embeddings for robust clustering and matching~\cite{DBLP:conf/mm/BredinG16,DBLP:conf/icassp/HooverCPSS18}. For example, Hoover \textit{et al.} \cite{DBLP:conf/icassp/HooverCPSS18} first detect speech segments and face tracks, and cluster them independently using pre-trained speech embeddings and  FaceNet~\cite{DBLP:conf/cvpr/SchroffKP15}. They then apply bipartite matching to assign each speech cluster to the face track cluster with maximum temporal overlap. However, unsupervised methods are not robust enough when the assumptions are not met or clustering results are noisy. \textit{Supervised} methods usually formulate ASD as a binary classification problem, and directly learn a frame-level binary classifier on fused audio-visual features \cite{DBLP:conf/icassp/RothCKMGKRSSXP20,DBLP:journals/corr/abs-1906-10555}. Some works~\cite{zhangmulti2019,DBLP:conf/cvpr/HuangK20} apply multi-task learning to boost performance, by introducing an additional contrastive loss on audio and visual features. An interesting recent work~\cite{DBLP:conf/cvpr/AlcazarCMPLAG20} constructs an Active Speaker Context ensemble to improve local predictions by querying adjacent time intervals and other speakers in the scene. Our work is a new and fundamentally different attempt in this direction, which proposes a methodical process to better incorporate different types of meaningful context information. We leverage such information in a unified framework to optimize all candidates jointly rather than independently for more robust ASD.
\vspace{-1ex}
\paragraph{Context-aware action recognition and detection} There has been a large body of work that focuses on how to integrate \textit{contextual} information in the field of action recognition and spatio-temporal action detection. For example, \cite{DBLP:conf/cvpr/GirdharCDZ19} uses a modified Transformer~\cite{DBLP:conf/nips/VaswaniSPUJGKP17} architecture to aggregate context from other people to recognize the target person's actions. \cite{DBLP:journals/corr/abs-2006-07976} proposes a high-order relation reasoning operator that enhances action recognition by modeling Actor-Context-Actor relations. \cite{DBLP:conf/eccv/WuKWZW20} improves spatio-temporal action detection by expanding actor bounding boxes to include more background context. We draw inspiration from these successful works and explore the use of contextual information in ASD.
\vspace{-1ex}
\paragraph{Gaze and social interaction} Evidence from psychology suggests that gaze direction plays an important role in social interactions~\cite{kendon1967some}, such as conversations. Since the VFoA of conversation participants typically converge on the active speaker, some previous work on speaker turn detection \cite{DBLP:conf/interspeech/JokinenHNY10} and audio-visual diarization of meetings \cite{DBLP:conf/mm/GarauBBO09,DBLP:conf/interspeech/GarauDB10,diaz2021audio} have reported good results by explicitly modeling head pose or gaze information with ad-hoc modules. In this work, we implicitly utilize such information through our spatial and relational contexts, which model the relationships among the candidates in an efficient and unified process.
\begin{figure*}
	\centering
	\includegraphics[width=\linewidth]{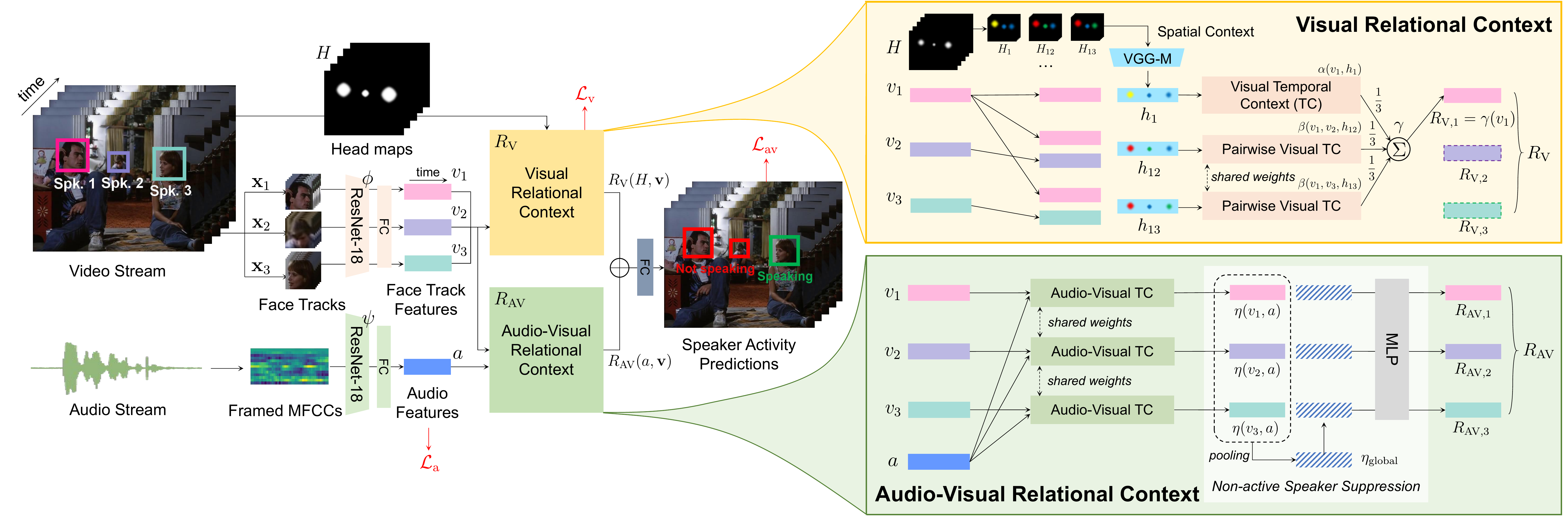}
	\vskip-2ex
	\caption{\textit{Model overview.} Given a video, we first extract face track features and audio features. Face scale and position information are encoded as 2D Gaussians and embedded with CNN layers, which we refer to as \textit{spatial context}. Next, we construct contextual audio-visual representations for each candidate, through intermediate \textit{visual relational context} and \textit{audio-visual relational context} modules. For the visual relational context, we introduce a permutation-equivariant layer to refine each speaker's visual representation by incorporating pairwise relationships and long-term \textit{temporal context}. For the audio-visual relational context, we model audio-visual affinity over time for each candidate and then suppress non-active speakers by contrasting their affinity features with others. The final refined visual and audio-visual representations are concatenated and passed through a shared prediction layer to estimate a confidence score for each visible candidate.}
	\label{fig:architecture}
	\vskip-2ex
\end{figure*}

\vspace{-2ex}

\section{The \NetName{} Model}
% high-level overview of UniCon
In this section, we describe our \NetName{} model, which integrates spatial, relational, and temporal context in a unified framework.
First, for each candidate, the scale and position of all candidates' faces are introduced as global \textit{spatial} context to complement facial information and help learn the relationships among the speakers. Each candidate is then contrasted with others from both a visual and audio-visual perspective in the \textit{relational} context modeling component. To further improve the robustness of the model's predictions, \textit{temporal} context is integrated. Finally, based on the aggregated contextual features, we generate speaker activity predictions for all speakers simultaneously with a shared prediction layer. Fig.~\ref{fig:architecture} provides an overview of our proposed approach.
\vspace{-2ex}
\subsection{Encoders}
Given an input video clip, we first crop face tracks for each candidate speaker and transform them into low-dimensional speaker descriptors for further analysis. Likewise, we encode each audio frame into a low-dimensional audio descriptor.
\vspace{-2ex}
\paragraph{Face Track Encoder:}
To encode short-term temporal dynamics, at each time step $t = 1,2,\dots, T$, where $T$ is the total number of time steps (i.e. frames), the $i$-th candidate's input ($i=1,2,\dots,N$ where $N$ is the number of visible candidates at time step $t$) is a stack of $k$ consecutive face crops $\mathbf x_i^t = \big(x_i^{t-[k/2]}, \dots, x_i^{t+[k/2]}\big)$ between step $t-[k/2]$ and $t+[k/2]$. Then an encoder $\phi$, which is a ResNet-18~\cite{DBLP:conf/cvpr/HeZRS16} is used to produce an average-pooled feature vector $\phi \big(\mathbf x_i^t\big)\in\mathbb{R}^{d}$ over the $k$ face crops. To keep the computational cost manageable, we choose $k=5$ for our experiments and reduce the dimensionality of each $\phi \big(\mathbf x_i^t\big)$ to $d'=128$ using a single shared fully-connected layer to obtain each candidate's final face track features:
\[v_i=\big(\mathrm{FC}\big(\phi\big(\mathbf x_i^1\big)\big),\mathrm{FC}\big(\phi\big(\mathbf x_i^2\big)\big),\dots,\mathrm{FC}\big(\phi\big(\mathbf x_i^T\big)\big)\in\mathbb{R}^{T\times d'}.\]
\paragraph{Audio Encoder:}
At each time step $t$ ($t=1,2,\dots, T$), we obtain audio representations from a $400$ms window preceding $t$. A ResNet-18 encoder $\psi$ takes $13$-dimensional MFCCs of the window as input and outputs a $512$-dimensional average-pooled feature. These features are also dimension-reduced to $128$ using a single fully-connected layer. We denote the final audio features by $a = (a_1, a_2,\dots, a_T)$.
\vspace{-2ex}
\subsection{Spatial Context}
The purpose of modeling spatial context is twofold. First, active speakers usually occupy a central position in the scene and depict higher visual saliency, especially in movies (commonly known as "the language of the lens"). Hence knowing the scale, position, and trajectory of a face in the video can help eliminate unlikely candidates. Second, people tend to look at the active speaker as they listen. Therefore, we wish to reflect such gaze-related information to some degree, by providing the model with both facial features and the relative positions of the candidates in the scene.

Specifically, we encode head positions of all candidates in the scene using $64\times 64$ coordinate-normalized maps of 2D Gaussians, which is motivated by \cite{DBLP:conf/cvpr/Marin-JimenezKM19}. The position and radius of each Gaussian represent the relative position and size of each candidate's face. Next, as shown in Fig.~\ref{fig:head-maps}, to further indicate the candidates' relationships, for each candidate $i$ ($i=1,2,\dots, N$) we construct his/her person-specific head map $H_i$ by generating a color-coded version of the initial Gaussian map in the following principle: yellow denotes candidate $i$, and blue denotes other candidates in the scene. To further facilitate the subsequent modeling of relationships between candidate $i$ and every other candidate $j$ ($j=1,2,\dots,N$, $j\ne i$), we tweak the color coding scheme to construct paired variants $H_{ij}$ as follows: red denotes candidate $i$, green denotes candidate $j$ and blue denotes the rest. Alternatively, $H_{ij}$ can be thought of an RGB image: the red channel shows the "subject", i.e. candidate $i$; the green channel shows the "object", i.e. candidate $j$; and the blue channel shows other "context" candidates, i.e. those other than $i$ and $j$ ($H_i$ is identified with $H_{ii}$). Then a VGG-M-inspired network with four 2D convolutional layers~\cite{DBLP:conf/cvpr/Marin-JimenezKM19} is used to embed these colored head maps $H_i$ and $H_{ij}$ into $64$-dimensional vectors per frame, $h_i$ and $h_{ij}$ for each candidate $i$ and candidate pair $(i, j)$ with $i\ne j$. We term the resulting embeddings $h_i$ and $h_{ij}$ candidate $i$'s \textbf{spatial context}.
\vspace{-4.5ex}
\begin{figure}[h]
	\centering
	\includegraphics[width=0.9\columnwidth]{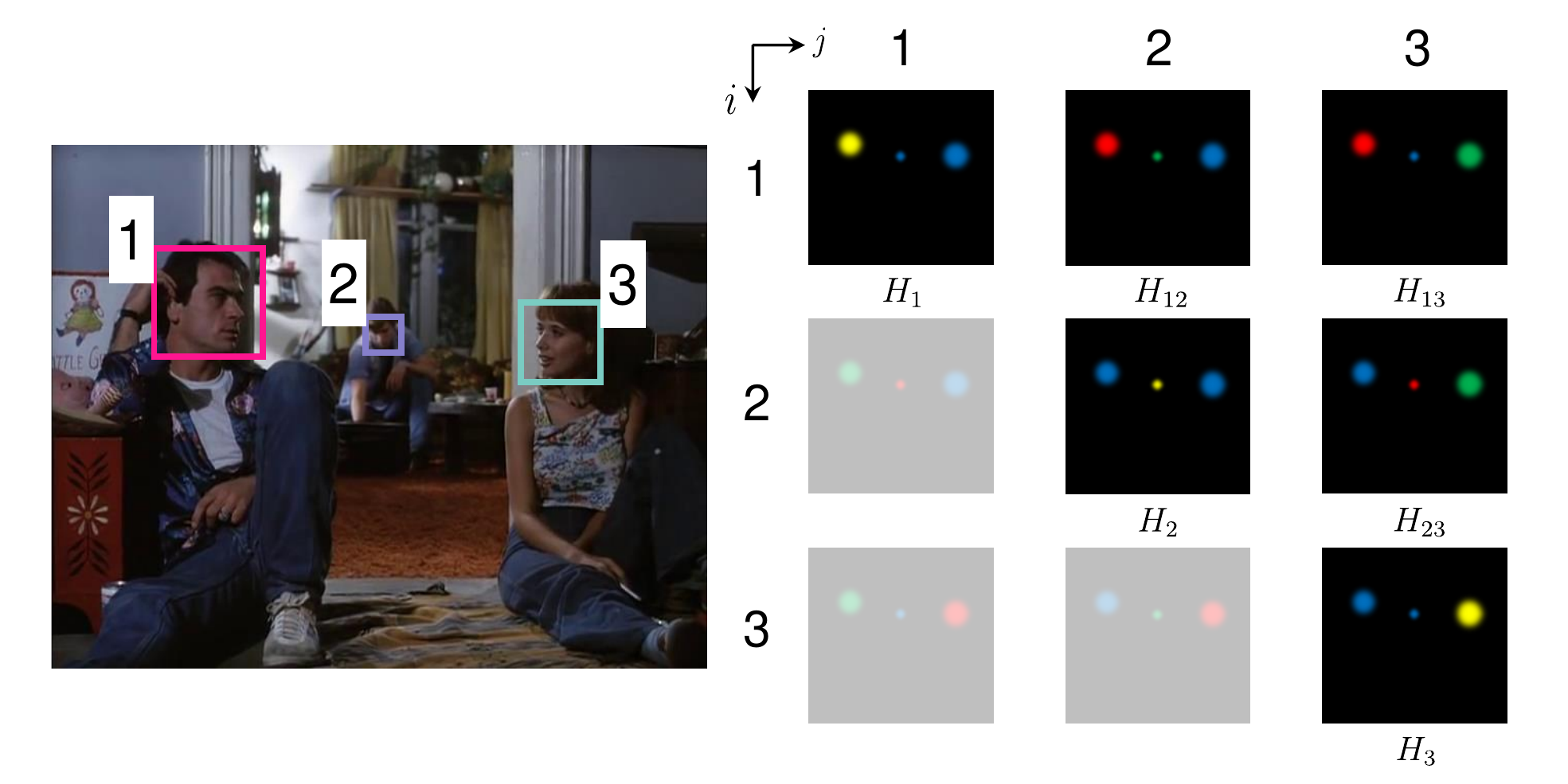}
	\vskip-3ex
	\caption{\textit{Head map construction.} For each candidate $i$, we construct his/her speaker-centric head map $H_i$ and pairwise head maps $\{H_{ij}\mid i\ne j\}$ for each candidate pair $(i, j)$. Note that the lower-triangular elements with $i > j$ are not used due to the skew-symmetric implementation (see Sec.~\ref{sec:relational-context}).}
	\label{fig:head-maps}
	\vskip-4.5ex
\end{figure}

\subsection{Relational Context}\label{sec:relational-context}
After obtaining face track and audio descriptors as well as the spatial context embeddings, we jointly model all candidates in the scene and refine each candidate's representations for robust ASD. Our motivation for modeling relational context is the inevitable presence of local ambiguities which make directly matching face motion (e.g. lip movements) and audio non-trivial. Therefore, we integrate different types of contextual clues and consider all candidates in the scene holistically to facilitate the ASD task. For example, a person who is speaking tends to receive more attention from others and is usually portrayed with higher visual saliency (especially in movies).

Specifically, our relational context component is designed to complete two natural sub-tasks for ASD: learning a contextual visual representation for \textit{visual voice activity detection}, and a contextual audio-visual representation for \textit{audio-visual affinity modeling}. The two resulting representations will then be fused for joint analysis to produce the final prediction. As shown in Fig.~\ref{fig:architecture}, our relational context contains two parts: \textbf{visual} ($R_\mathrm{V}$) and \textbf{audio-visual relational context} ($R_\mathrm{AV}$). This special decoupled design strengthens the model's robustness to synchronization errors between audio and video. We now elaborate on the design of $R_\mathrm{V}$ and $R_\mathrm{AV}$.
\paragraph{Visual Relational Context:}
\iffalse
When jointly optimizing all candidates in the scene, the order in which the candidates are arranged within the context should be immaterial. To simultaneously process all candidates, we introduce a \textit{permutation-equivariant} layer $\gamma$ which produces exactly one output per candidate, while preserving the order they are provided to the network. In other words, given any permutation of the $N$ candidate speakers $\sigma\in S_N$, where $S_N$ is the symmetry group in $N$ letters, the following holds:
\begin{equation}
	\sigma\big(\gamma(v_1),\dots,\gamma(v_N)\big) = \big(\gamma\big(v_{\sigma(1)}\big),\dots,\gamma\big(v_{\sigma(N)}\big)\big).
\end{equation}
\fi
Our key idea is to represent each candidate's visual activity by aggregating his/her locally perceived activity and all pairwise interactions with other candidates in the scene. Specifically, we design a \textit{permutation-equivariant} layer $\gamma$ to process all candidates simultaneously, while preserving the order in which they are provided to the network. Denote the input visual feature stack by $\mathbf v=(v_1,v_2,\dots, v_N)\in\mathbb{R}^{N\times T\times d'}$. For each candidate $i=1,2,\dots,N$, the output in the $i$-th position of $R_\mathrm{V}(\mathbf v)$ is
\begin{equation}\label{eq:equivariant-gamma}
	R_{\mathrm{V},i} = \gamma(v_i) = \frac 1N\bigg[\alpha(v_i, h_i) + \sum_{j\ne i}\beta(v_i, v_j, h_{ij})\bigg],
\end{equation}
where $\alpha(\cdot,\cdot)$ and $\beta(\cdot, \cdot, \cdot)$ are two networks that respectively model visual activity and pairwise visual interactions, $v_i$ denotes candidate $i$'s visual features, $h_i$ is the head map embedding for candidate $i$, and $h_{ij}$ is the head map embedding for the candidate pair $(i, j)$.

Here, the parameters of $\alpha$ are shared across all candidates, and the parameters of $\beta$ are shared across all candidate pairs, so Eq.~\eqref{eq:equivariant-gamma} is directly applicable to any number of co-occurring candidates $N$. This in turn means that our network can analyze a theoretically unlimited number of candidates at test time. On the surface, Eq.~\eqref{eq:equivariant-gamma} appears to suggest that obtaining $R_\mathrm{V}(\mathbf v)$ requires $N^2+N$ evaluations in total. In our implementation, we reduce this number by half to $N(N+1)/2$ by parameterizing $\beta(\cdot,\cdot, \cdot)$ as a \textit{skew-symmetric} function:%, that is,
\begin{equation}\label{eq:skew-symmetry}
	\beta_{ji}=\beta(v_j,v_i,h_{ji}) = -\beta(v_i,v_j,h_{ij})=-\beta_{ij}.
\end{equation}
Our intuition is that $\beta_{ij}$ should learn to represent the amount of attention that candidate $i$ directs at candidate $j$, and rank the relative ``activeness" and visual saliency of candidate $i$ against candidate $j$. In this case, it suffices to express the direction of such relationships using a single positive or negative sign for each component.
\vspace{-1.5ex}
\paragraph{Audio-Visual Relational Context:}
We model the affinity between the audio and each speaker's face track in terms of both synchrony and cross-modal biometrics. We fuse per-frame face track features with the audio features via concatenation, and then pass them through a shared network $\eta$ to compute local A-V affinity features $\eta(v_i, a)$. Considering that local A-V affinity estimation is vulnerable to signal ambiguity (e.g. low-resolution faces, profile faces, noisy audio etc.), we aggregate evidence from all candidates to make more reliable predictions: if one or some candidates display strong A-V agreement, then the network can accordingly lower its affinity predictions for other candidates to mitigate false positives. We therefore design a learnable module to adaptively suppress only the non-active candidates whose A-V affinities are of significantly lower magnitudes, while leaving the active candidates' features "as-is". To this end, we introduce an element-wise max-pooling operation over all A-V affinity features to obtain a global representation. Specifically, the pooled feature is computed as $\eta_{\mathrm{global}} = \max_{1\le i\le N}\eta(v_i, a)$, 
\iffalse
\begin{equation}\label{eq:eta_global}
	\eta_{\mathrm{global}} = \max_{1\le i\le N}\eta(v_i, a),
\end{equation}
\fi
and then concatenated to each candidate's initial features $\eta(v_i, a)$. The concatenated features are further processed with two fully connected layers to generate each candidate's final contextual audio-visual representation $R_{\mathrm{AV},i}$. We name this operation \textit{non-active speaker suppression} after its effect.
\vspace{-2ex}
\subsection{Temporal Context}\label{sec:temporal-context}
As shown in Fig.~\ref{fig:architecture}, we incorporate temporal context in the networks $\alpha$, $\beta$, and $\eta$ inside $R_\mathrm{V}$ and $R_\mathrm{AV}$. This benefits ASD in two ways: first, it improves the consistency of the relational context modeling process and smoothes out local, instantaneous noises; second, it helps to alleviate synchronization errors between the audio and the video stream, an issue that is ubiquitous with in-the-wild videos.

We choose two basic architectures as our temporal modeling back-end: temporal convolutions (1D CNNs), and bi-directional Gated Recurrent Units (Bi-GRUs). The former is favorable since it has a fixed temporal receptive field, and can be easily adapted for online settings. The latter is more reliable since it has access to both past and future information. Hence, for the rest of the paper, by mentioning \textbf{temporal context} we refer to the usage of a one-layer Bi-GRU backend with $256$ cells. Otherwise, we apply a stack of two 1D convolutional layers of kernel size $3$ interleaved with Batch Normalization~\cite{DBLP:conf/icml/IoffeS15} and ReLU activation by default.

Finally, the refined contextual representations $R_\mathrm{V}$ and $R_\mathrm{AV}$ are concatenated and fed to a fully-connected layer that is shared across candidates. The outputs are passed through sigmoid activation, resulting in real values between $0$ and $1$ that indicate each candidate's probability of being the active speaker.
\vspace{-2ex}
\subsection{Losses}\label{sec:losses}
The model is trained end-to-end with a multi-task loss formulation. Since the goal is to predict a binary speaking/not speaking label, for each loss term we apply the standard binary cross-entropy (BCE) loss, averaged over all time steps. The BCE loss is defined as
\begin{equation}
	\mathcal{L}_{\mathrm{BCE}}(y, \hat y) = -\hat y\log y - (1-\hat y)\log(1-y),
\end{equation}
where $y$ and $\hat y$ are the predictions and ground truth labels, respectively. For a scene with $T$ frames and $N$ candidates, let $A_{\mathrm{aux}}$ and $V_{\mathrm{aux}}$ denote the audio and visual auxiliary prediction layer, $AV_{\mathrm{pred}}$ the final audio-visual prediction layer, and $\hat{y}_{\mathrm{v},i}^{1,2,\dots,T}$ and $\hat{y}_{\mathrm{av},i}^{1,2,\dots,T}$ the corresponding visual and audio-visual ground truths of the $i$-th candidate. We add two auxiliary losses $\mathcal{L}_\mathrm{a}$ and $\mathcal{L}_\mathrm{v}$ to learn discriminative features for both visual and audio streams. The \textit{audio} prediction loss is
\begin{equation}
	\mathcal{L}_\mathrm{a} = \frac 1T\sum_{t=1}^T \mathcal{L}_{\mathrm{BCE}}\big(\sigma(A_{\text{aux}}(a^t)), \hat{y}_{\mathrm a}^t\big),
\end{equation}
where $\hat{y}_{\mathrm a}^t = \max_{1\le i\le N}\hat{y}_{\mathrm{av},i}^t$ is the audio ground truth indicating whether at least one of the candidates is speaking, and $\sigma(\cdot)$ is the sigmoid function. %When training with the relational context component on multiple candidates, we drop the $\mathcal{L}_\text{a}$ term as we observe serious overfitting in the audio modality at this stage.

When training on a single candidate (\textit{i.e.} without employing relational context), 
the \textit{visual} prediction loss and main \textit{audio-visual} prediction loss are given by 
\begin{align}
	\begin{split}
	\mathcal{L}_\text{v} &= \frac 1T\sum_{t=1}^T \mathcal{L}_{\mathrm{BCE}}\big(\sigma(V_{\text{aux}}(v_i^t)), \hat{y}_{\mathrm v,i}^t\big),\\
	\mathcal{L}_\text{av} &= \frac 1T\sum_{t=1}^T \mathcal{L}_{\mathrm{BCE}}\bigg(\sigma\big(AV_{\text{pred}}\big(a^t\oplus v_i^t\big)\big), \hat{y}_{\mathrm{av},i}^t\bigg),
	\end{split}
\end{align}
where $\oplus$ denotes feature concatenation. When training with relational context on multiple candidates, the losses are slightly different. To compute the visual and audio-visual losses, we replace the inputs to the auxiliary prediction layers with (concatenated) contextual representations, and aggregate losses from all candidates:
\begin{align}\label{eq:loss-nspk}
\begin{split}
	\mathcal{L}_\text{v} &= \frac 1N\sum_{i=1}^N\frac 1T\sum_{t=1}^T \mathcal{L}_{\mathrm{BCE}}\bigg(\sigma\big(V_{\text{aux}}\big(R_{\mathrm{V}, i}^t\big)\big), \hat{y}_{\mathrm v,i}^t\bigg),\\
	\mathcal{L}_\text{av} &= \frac 1N\sum_{i=1}^N\frac 1T\sum_{t=1}^T \mathcal{L}_{\mathrm{BCE}}\bigg(\sigma\big(AV_{\text{pred}}\big(R_{\mathrm{V}, i}^t\oplus R_{\mathrm{AV}, i}^t\big)\big), \hat{y}_{\mathrm{av},i}^t\bigg).
\end{split}
\end{align}

Finally, the \textit{total} loss is defined as follows:
\begin{equation}\label{eq:total_loss}
	\mathcal{L} = \mathcal{L}_\text{a} + \mathcal{L}_\text{v} + \mathcal{L}_\text{av}.
\end{equation}

Note that in practice, $N$ may vary over time as candidates enter and leave the scene.
\vspace{-1.5ex}

\section{Experiments}
To validate the effectiveness of our model, we thoroughly evaluate and analyze the performance of \NetName{} from two aspects. First, we perform comprehensive ablation studies and comparison on a popular and challenging ASD benchmark derived from diverse, in-the-wild movies, AVA-ActiveSpeaker. Second, we perform cross-dataset testing on three additional datasets, Columbia, RealVAD, and AVDIAR. Example frames from each dataset are shown in Fig.~\ref{fig:dataset-teaser}.

	\begin{figure}
		\centering
		\subfigure[AVA-ActiveSpeaker]{\includegraphics[width=0.45\columnwidth]{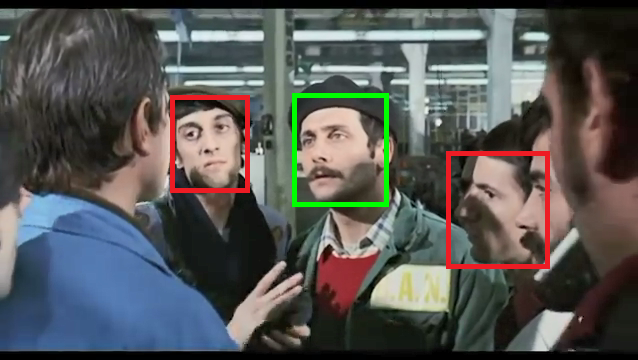}}
		\hfill
		\subfigure[Columbia]{\includegraphics[width=0.45\columnwidth]{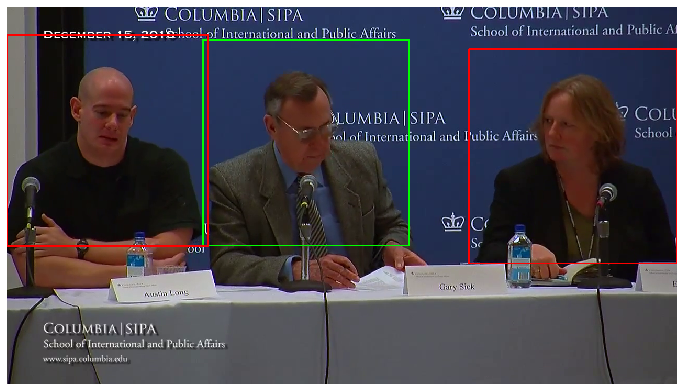}}\\[-6pt]
		\subfigure[RealVAD]{\includegraphics[width=0.45\columnwidth]{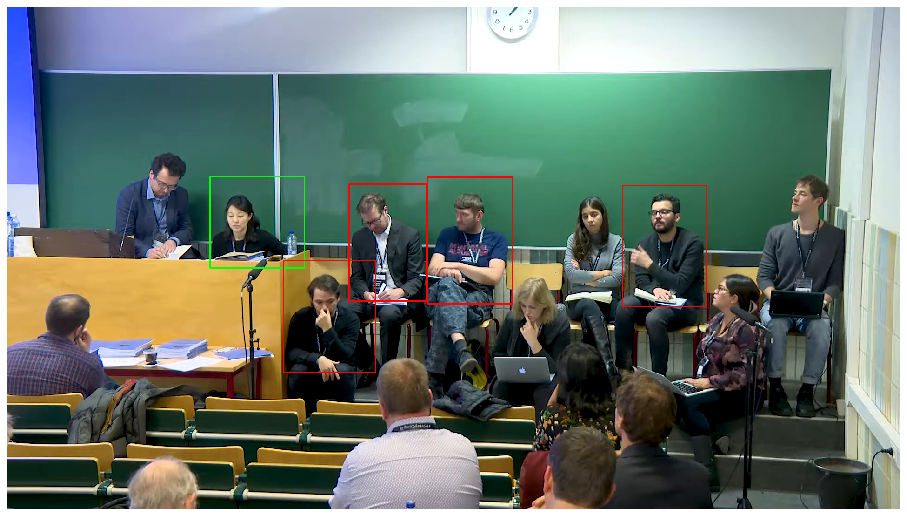}}
		\hfill
		\subfigure[AVDIAR]{\includegraphics[width=0.4\columnwidth]{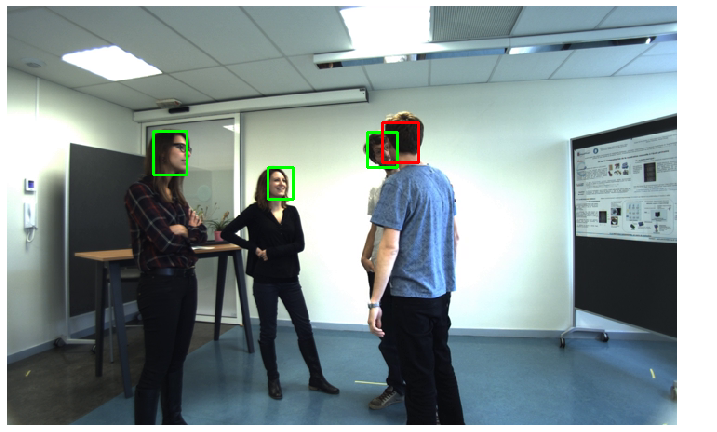}}
		\vskip-4ex
		\caption{\textit{Example frames from the datasets used in this paper.} Green boxes denote the active speakers.}
		\label{fig:dataset-teaser}
		\vskip-4ex
	\end{figure}

\vspace{-2.5ex}
\subsection{Experimental Setup}\label{sec:setup}
\paragraph{Datasets:} The \textbf{AVA-ActiveSpeaker} dataset \cite{DBLP:conf/icassp/RothCKMGKRSSXP20} consists of $262$ YouTube movies from film industries around the world. Each video is annotated from minutes $15$ to $30$. The dataset provides face bounding boxes that are linked into face tracks and labeled for both speech activity and whether the speech is audible. As shown in Fig.~\ref{fig:dataset-teaser}a, the dataset is highly challenging as it contains occlusions, low-resolution faces, low-quality audio, and varied lighting conditions. It has become a mainstream benchmark for the ASD task.

The \textbf{Columbia} dataset \cite{DBLP:conf/eccv/ChakravartyT16} annotates $35$ minutes of a one-hour panel discussion video featuring $6$ unique panelists (Bell, Bollinger, Lieberman, Long, Sick, Abbas). The panelists directly face the camera, but occasionally look elsewhere, or perform spontaneous activities (e.g. drinking water). Upper body bounding boxes and voice activity ground truth labels are provided, as shown in Fig.~\ref{fig:dataset-teaser}b.

The \textbf{RealVAD} dataset~\cite{9133504} (Fig.~\ref{fig:dataset-teaser}c) is also constructed from a panel discussion lasting approximately $83$ minutes. The video is recorded using a static, mounted camera, capturing the $9$ panelists in a full shot. The panelists sit in two rows, and at any given time can be looking anywhere. In addition, the panelists perform natural, spontaneous actions that may result in partial occlusion of the face and mouth (e.g. touching faces, cupping chins). Similar to Columbia, upper body bounding boxes and voice activity labels are provided. These two datasets are often adopted by works on visual voice activity detection (V-VAD) and ASD~\cite{DBLP:conf/iccvw/ShahidBM19,Shahid_2021_WACV,DBLP:conf/accv/ChungZ16a,DBLP:conf/eccv/AfourasOCZ20}.

The \textbf{AVDIAR} dataset \cite{DBLP:journals/pami/GebruBLH18} (Fig.~\ref{fig:dataset-teaser}d) provides $27$ recordings of informal conversations involving one to four participants. The dataset is extremely challenging due to varying levels of speaker movement and speech overlap, occlusions, profile faces, and even back-facing speakers. The dataset was originally used to evaluate speaker diarization and tracking systems. We re-purpose the dataset to evaluate our ASD model by aggregating per-frame ASD results into diarization results.
\vspace{-2ex}
\paragraph{Data Preprocessing:} We preprocess each dataset by cropping the face tracks of each visible candidate. For the \textbf{AVA-ActiveSpeaker} dataset, to group face tracks into ``scenes" for training and testing, each video is first segmented into shots with \texttt{ffmpeg}~\cite{DBLP:journals/ijig/Lienhart01}. Within each shot, disconnected tracks belonging to the same candidate are then merged based on bounding box overlap with an IoU threshold of $0.8$. All videos are resampled to $25$fps, and ground truth labels are computed according to frame timestamps via nearest-neighbor interpolation. The \textbf{Columbia} and \textbf{RealVAD} datasets do not provide face bounding boxes, so face detection is performed with an off-the-shelf RetinaFace detector~\cite{DBLP:conf/cvpr/DengGVKZ20}. The resulting bounding boxes are expanded by $1.3$x to mimic AVA-style detections. For the \textbf{AVDIAR} dataset, the provided bounding boxes are used directly.
\vspace{-1.5ex}
\paragraph{Train/test Split and Evaluation Metric:}
For \textbf{AVA-ActiveSpeaker}, the official train/validation/test split is adopted. The test set is held out for the ActivityNet challenge and unavailable, so we perform our analysis on the validation set instead, as several previous methods do~\cite{DBLP:conf/cvpr/AlcazarCMPLAG20,zhangmulti2019,DBLP:journals/corr/abs-1906-10555}. The \textbf{Columbia} and \textbf{AVDIAR} datasets do not provide training sets, so we only perform zero-shot testing. For Columbia, following previous practice \cite{DBLP:conf/eccv/ChakravartyT16,DBLP:conf/accv/ChungZ16a}, we report results on all but one speaker (Abbas) which is usually held out for validation. For \textbf{RealVAD}, in addition to zero-shot results, we also report leave-one-out cross-validation results for each of the $9$ panelists to compare with previous work. We only train on scenes in which the test speaker does not co-occur with the training speakers. 
%We reserve the last $10\%$ of each training scene for validation.

Finally, we report the metric that was commonly adopted by previous work for each dataset: mean Average Precision (mAP) and Area Under Receiver Operating Characteristic Curve (AUROC) for AVA-ActiveSpeaker, frame-wise $F$-1 score for Columbia and RealVAD, and Diarization Error Rate (DER) for AVDIAR, which is defined as the sum of false alarm rate, missed speech rate, and speaker confusion rate.
\vspace{-2ex}
\paragraph{Implementation Details:} We implement our model with PyTorch and the \texttt{pytorch-lightning} package. All models are trained using the AdamW optimizer~\cite{DBLP:conf/iclr/LoshchilovH19}. The network parameters are initialized using He initialization~\cite{DBLP:conf/iccv/HeZRS15}. During training, we augment the data via random horizontal flipping and uniform corner cropping along the input face tracks, followed by random adjustments to brightness, contrast, and saturation. All cropped face tracks are resized to $144\times 144$, and randomly cropped to $128\times 128$ for training. We use a central $128\times 128$ patch for testing. We only run one trial with a fixed random seed for all experiments to ensure the results are comparable, but find the results to be stable under different network initializations and data augmentations. During inference, occasional long segments that do not entirely fit into GPU memory are split into shorter, fixed-size chunks and the predictions are re-combined later.

To facilitate training, we apply curriculum learning~\cite{DBLP:conf/icml/BengioLCW09} by first training a single-candidate model without relational context according to Eq.~\eqref{eq:total_loss}, and then continuing to train on up to $3$ candidates, \textit{i.e.} $N=3$ in Eq.~\eqref{eq:equivariant-gamma}, but without the $\mathcal{L}_{\mathrm{a}}$ term in Eq.~\eqref{eq:total_loss}. This is theoretically and empirically enough to learn the relevant parameters since $R_\mathrm{V}$ captures \textit{pairwise} interactions, and we observe diminishing returns by sampling $4$ or more candidates (note that we can still test on $N>3$ candidates due to parameter sharing). In addition, we follow the sampling strategy in \cite{DBLP:conf/cvpr/AlcazarCMPLAG20} during training, and randomly sample a $1.12$s segment ($28$ frames at $25$fps)\footnote{The mean speech segment duration is $1.11$s on the AVA-ActiveSpeaker dataset~\cite{DBLP:conf/icassp/RothCKMGKRSSXP20}.} from every training example in the dataset for each epoch. Therefore, our epoch size correlates with the number of identities or scenes rather than face detections, which prevents over-fitting.
\vspace{-2ex}
\subsection{Ablation Studies}\label{sec:ablations}
In this subsection, we provide a thorough ablation study on the modeling components on the challenging AVA-ActiveSpeaker dataset.
\vspace{-4ex}
\paragraph{Spatial Context:} We compare two different ways to incorporate spatial context. The first approach applies \textit{early fusion} and concatenates speaker-centric spatial embeddings $h_i$ with the $i$-th candidate's face track features before dimensionality reduction. In contrast, the second approach applies \textit{late fusion} and concatenates $h_i$ with the dimension-reduced face track features $v_i$. As shown in Table~\ref{table:head-maps}, both improve upon the simplest baseline that only uses convolutional temporal context (see Sec.~\ref{sec:temporal-context}) on audio and face track features, but early fusion performs $0.8$\% worse than late fusion, likely because information is lost too early as features pass the $128$-dim bottleneck. Thus, we adopt late fusion hereafter. As shown in Table~\ref{table:ablations}, simply concatenating the spatial context embeddings to the face track features (denoted by \textbf{+S}) already yields stronger visual representations that improve the baseline ($+1.7$\% mAP).
\vspace{-2.5ex}
\begin{table}[h]
	\caption{\textit{Comparison w.r.t. the stage at which spatial context is introduced.} We observe that applying late fusion yields better results since it helps to retain more spatial information for subsequent modeling.}
	\vskip-3ex
	\label{table:head-maps}
	\begin{center}
		\begin{tabular}{l|c|c}
			\toprule
			\textbf{Method} & \textbf{mAP (\%)} & \textbf{AUROC} \\
			\midrule
			% version_106
			Baseline & $84.0$ & $0.93$ \\
			% version_107
			% mAP 0.8330118795487191 roc_auc 0.9265507949283783
			% mAP 0.8504181198406724 roc_auc 0.9345415355416297
			Early fusion & $85.0$ & $0.93$ \\
			% version_109
			\textbf{Late fusion} & $\bf 85.7$ & $\bf 0.94$ \\
			\bottomrule
		\end{tabular}
	\end{center}
	\vskip-3.5ex
\end{table}

\vspace{-1ex}
%This is in line with our intuition: background characters are easily identified through their relative sizes and motion patterns, which are different from main actors.
\paragraph{Relational Context:}
We assess the efficacy of our relational context component. As shown in Table~\ref{table:ablations}, incorporating the relational context module yields noticeable improvements of $1.7$\% absolute over the 1D CNN baseline when used alone (\textbf{+R}), and works in synergy with spatial context ($+2.6$\% mAP for \textbf{+S+R}). We observe that this improvement is especially pronounced in multi-speaker and small face scenarios. This is because \NetName{} can model the relationships among the candidates in the scene, which leads to a better holistic understanding under these challenging settings. Moreover, we introduce weight sharing strategies within a permutation-equivariant formulation, which enables our model to process a theoretically unlimited number of speakers at test time.
\vspace{-1.5ex}
\paragraph{Temporal Context:}
Finally, incorporating temporal context with Bi-GRUs yields the most substantial improvement, which is reasonable because Bi-GRUs have access to each candidate's complete track history, and can thus make more reliable predictions. By leveraging long-term temporal context, our model ultimately achieves $92.0$\% mAP, improving upon our initial baseline by as much as $8.0$\%.
\begin{table}
	\centering
	\vskip-3ex
	\caption{\textit{Ablation results on the AVA-ActiveSpeaker validation set.} \textbf{S}: Spatial Context; \textbf{R}: Relational Context; \textbf{T}: Temporal Context. Predictions from models without temporal context are smoothed over $11$-frame windows via Wiener filtering.}
	\vskip-2.5ex
	\label{table:ablations}
	\resizebox{0.85\columnwidth}{!}{
		\begin{tabular}{l|c|c|c}
			\toprule
			\textbf{Method} & \textbf{\#Params} & \textbf{mAP (\%)} & \textbf{AUROC}\\
			\midrule
			% version_106
			% mAP 0.8224600256899731 roc_auc 0.9205524247843067
			% mAP 0.8401552380444571 roc_auc 0.9293347368213922
			Baseline & $23.2$M & $84.0$ & $0.93$ \\
			% version_109
			% mAP 0.841241917173582 roc_auc 0.9296624801256668
			% mAP 0.8573910355141074 roc_auc 0.9375171326156119
			+S & $23.4$M & $85.7$ & $0.94$ \\
			% mAP 0.8385913434424734 roc_auc 0.9279589330057654
			% mAP 0.8569203189558342 roc_auc 0.9371191010728116
			+R & $23.8$M & $85.7$ & $0.94$ \\
			% version_115
			% mAP 0.8507699834089036 roc_auc 0.9332089598782902
			% mAP 0.8657510674959639 roc_auc 0.9405652472504447
			+S+R & $24.0$M & $86.6$ & $0.94$ \\
			% version_146
			% mAP 0.8964767460084616 roc_auc 0.9545367983013869
			\midrule
			+T (Bi-GRU) & $23.0$M & $89.6$ & $0.95$ \\
			% version_101
			% mAP 0.9033484862842304 roc_auc 0.957823575408724
			+S+T & $23.2$M & $90.3$ & $0.96$ \\
			% version_157
			% mAP 0.9040215606508509 roc_auc 0.9581817442375401
			+R+T & $23.5$M & $90.4$ & $0.96$ \\
			% version_80
			% mAP 0.9196214010080134 roc_auc 0.9656457503644401
			\textbf{+S+R+T (\NetName{})} & $23.8$M & $\bf 92.0$ & $\bf 0.97$ \\
			\bottomrule
		\end{tabular}
	}
	\vskip-2.5ex
\end{table}

\vspace{-4ex}
\paragraph{Non-active Speaker Suppression:}
In this part, we discuss the role of non-active speaker suppression in ASD. As shown in Table~\ref{table:suppression}, we compare two ways of integrating global information against a baseline that does not apply non-active speaker suppression. One uses mean-pooling to obtain $\eta_{\mathrm{global}}$, and the other uses max-pooling. Both improve over the model that only introduces visual relational context, and max-pooling performs slightly better, yielding fewer false negatives. One possibility is that max-pooling back-propagates more gradients to the most salient speaker, while mean-pooling results in weaker gradients for the true active speaker, making the model less confident.

\begin{table}
	\caption{Effect of implementing non-active speaker suppression with different pooling methods.}
	\vskip-3ex
	\label{table:suppression}
	\begin{center}
		\begin{tabular}{l|c|c}
			\toprule
			\textbf{Pooling}             & \textbf{mAP (\%)} & \textbf{AUROC}  \\
			\midrule
			% version_142
			% mAP 0.9079781855686699 roc_auc 0.959520864572852
			None (no suppression) & $90.8$ & $0.96$ \\
			% version_143
			% mAP 0.9159581583383972 roc_auc 0.9635663469449117
			Mean-pooling & $91.6$ & $0.96$ \\
			% version_115
			\textbf{Max-pooling} & $\bf 92.0$ & $\bf 0.97$ \\
			\bottomrule
		\end{tabular}
	\end{center}
	\vskip-4ex
\end{table}

\vspace{-2ex}
\subsection{Comparison with the State-of-the-Art}
As shown in Table~\ref{table:ava-sota}, our full \NetName{} outperforms previous audio-visual ASD methods including the state-of-the-art by a large margin. Remarkably, we achieve $92.0$\% mAP without any pre-training, surpassing $90$\% for the first time on the AVA-ActiveSpeaker validation set at the time of submission. This overall performance strongly supports the effectiveness and superiority of our \NetName{}. It is also worth noting that the \textbf{+S+T} model in Table~\ref{table:ablations} provides a very strong baseline ($90.3$\% mAP) that already exceeds all methods available for comparison here, which further confirms the benefits of our model. As a side note, applying ImageNet pre-training to the encoders further boosts performance by $0.2$\% to $92.2$\%.
\vspace{-2.5ex}
\begin{table}[h]
	\centering
	\caption{\textit{Comparison with previous work on the AVA-ActiveSpeaker validation set.} For each method, we copy the results from its original paper. mAP is calculated using the official evaluation tool, after interpolating our predictions to the timestamps in the original annotations.}
	\label{table:ava-sota}
	\vskip-2.5ex
	\resizebox{0.9\columnwidth}{!}{
		\begin{tabular}{l|c|c|c}
			\toprule
			\textbf{Method} & \textbf{Pre-training?}  & \textbf{mAP (\%)} & \textbf{AUROC}\\
			\midrule
			AV-GRU \cite{DBLP:conf/icassp/RothCKMGKRSSXP20} & $\times$
			& - & $0.92$\\ 
			Optical Flow \cite{DBLP:conf/cvpr/HuangK20} & $\times$
			& - & $0.93$\\ 
			Multi-Task \cite{zhangmulti2019} & \checkmark
			& $84.0$& -\\
			VGG-LSTM \cite{DBLP:journals/corr/abs-1906-10555} & \checkmark
			& $85.1$ & -\\ 
			VGG-TempConv \cite{DBLP:journals/corr/abs-1906-10555} & \checkmark
			& $85.5$ & -\\ 
			ASC \cite{DBLP:conf/cvpr/AlcazarCMPLAG20} & \checkmark
			& $87.1$ & -\\ 
			\textbf{Ours (\NetName{})} & $\times$
			& $\bf 92.0$ & $\bf 0.97$\\
			\textbf{Ours (\NetName{})} & \checkmark & $\bf 92.2$ & $\bf 0.97$ \\
			\bottomrule
		\end{tabular}
	}
	\vskip-3.5ex
\end{table}

\vspace{-1ex}
\subsection{Cross-dataset Evaluation}\label{sec:cross-dataset-eval}
Apart from evaluating on the large-scale AVA-ActiveSpeaker dataset, we also conduct cross-dataset evaluation on three other datasets: Columbia, RealVAD and AVDIAR. These datasets contain challenges that are previously rare, if not unseen in AVA-ActiveSpeaker, such as overlapped speech, reverberation, extreme poses, heavy face occlusion, and different spatial distributions of candidates. Our cross-dataset evaluation protocols include zero-shot testing on all three datasets using the model trained on AVA-ActiveSpeaker, and leave-one-out testing after fine-tuning on RealVAD. We report both visual-only and audio-visual performance. The former is obtained using only the auxiliary visual prediction layer $V_{\mathrm{aux}}$ in Sec.~\ref{sec:losses}.
\vspace{-1.5ex}
\paragraph{Zero-shot Testing:}
Overall, our \NetName{} model achieves good zero-shot generalization performance on all three datasets. In particular, it achieves state-of-the-art performance on both Columbia ($90.9\%$ average $F$-1 score) and RealVAD ($80.30\%$ average $F$-1 score). On AVDIAR, we are the first to report diarization performance using single-channel audio and a single camera viewpoint.

On the \textbf{Columbia} dataset, 
we outperform all previous state-of-the-art under both visual-only and audio-visual settings. It is worth noting that the listed visual-only methods~\cite{DBLP:conf/eccv/ChakravartyT16,DBLP:conf/iccvw/ShahidBM19,Shahid_2021_WACV} use the entire upper body, while we only use cropped face tracks. We also outperform previous audio-visual models~\cite{DBLP:conf/eccv/AfourasOCZ20,DBLP:conf/accv/ChungZ16a} that are pre-trained on datasets a magnitude larger ($224$ and $606$ hours, respectively).
% Finally, while previous methods choose the decision threshold based on the equal error rate (EER) of the held-out speaker, we find that such practice actually degrades our \NetName{} model's performance. We believe this is because our model makes adaptive decisions by considering scene context, and setting a threshold based on data from a particular speaker becomes counterproductive. For such reasons, we apply a fixed decision threshold of $0.5$ for all cross-dataset evaluations.
Similarly, on \textbf{RealVAD}, our zero-shot visual-only and audio-visual results outperform the only existing method available for equal comparison by a large margin ($87.22$\% V, $80.30$\% AV vs. $53.04$\%). An interesting observation is that on RealVAD and Columbia, our visual-only model outperforms the audio-visual model in most cases. This is because labels provided with the two datasets are not frame-accurate; our audio-visual model successfully detects short inter-sentence pauses that are not annotated as "not speaking" by the dataset curators. Due to the page limit, please refer to the supplementary video for details.

On \textbf{AVDIAR}, our model provides the first baseline 
that only uses mono microphone input, and a single camera viewpoint -- a highly challenging setup. Nevertheless, our audio-visual model still achieves $55.16\%$ DER on average, which is a very promising zero-shot result. Moreover, our full \NetName{} model reduces DER by $4$ to $5$\% over the \textbf{+S+T} model that does not model relationships among the candidates, which once again supports the efficacy of our relational context in multi-speaker scenarios.
\vspace{-2ex}
\paragraph{Fine-tuned Results:} We also evaluate fine-tuned leave-one-out performance on RealVAD (Columbia and AVDIAR do not provide training splits). After 
fine-tuning on homogeneous data, our model quickly adapts to the unseen scenario (panel discussion) and noises (e.g. spontaneous actions near the face). The average audio-visual $F$-1 score is improved by $5.34$\% from $80.30$\% to $85.64$\%, and the average visual $F$-1 score is improved by $1.47$\% from $87.22$\% to $88.69$\%. More detailed results can be found in the appendix.
\vspace{-1.5ex}
\subsection{Performance Breakdown}
In this part, we provide a more in-depth analysis of three known challenging scenarios on AVA-ActiveSpeaker. We provide a comparison with the existing state-of-the-art and highlight some advantages and appealing properties of the \NetName{} model.
\vspace{-1ex}
\paragraph{Low-resolution Faces.} We summarize performance for different face sizes in Fig.~\ref{fig:breakdown}a. Following previous evaluation procedures \cite{DBLP:conf/icassp/RothCKMGKRSSXP20,DBLP:conf/cvpr/AlcazarCMPLAG20}, we partition the dataset into three bins by the widths of the detected faces: small (width $\le$ 64 pixels), medium (width between $64$ and $128$ pixels), and large (width $>128$ pixels). While the \textbf{+S+T} model provides a strong baseline with $90.3$\%mAP that already outperforms the previous state-of-the-art \cite{DBLP:conf/cvpr/AlcazarCMPLAG20}, our full model which exploits relational context yields further performance gains of around $2$\% mAP absolute on all subsets. In particular, on the ``small" subset, we outperform the previous state-of-the-art ($56.2$\% mAP) by a large margin of $15.1$\%.

\begin{figure}
	\vskip2ex
	\subfigure[mAP breakdown by face resolution.]{\includegraphics[width=0.48\columnwidth]{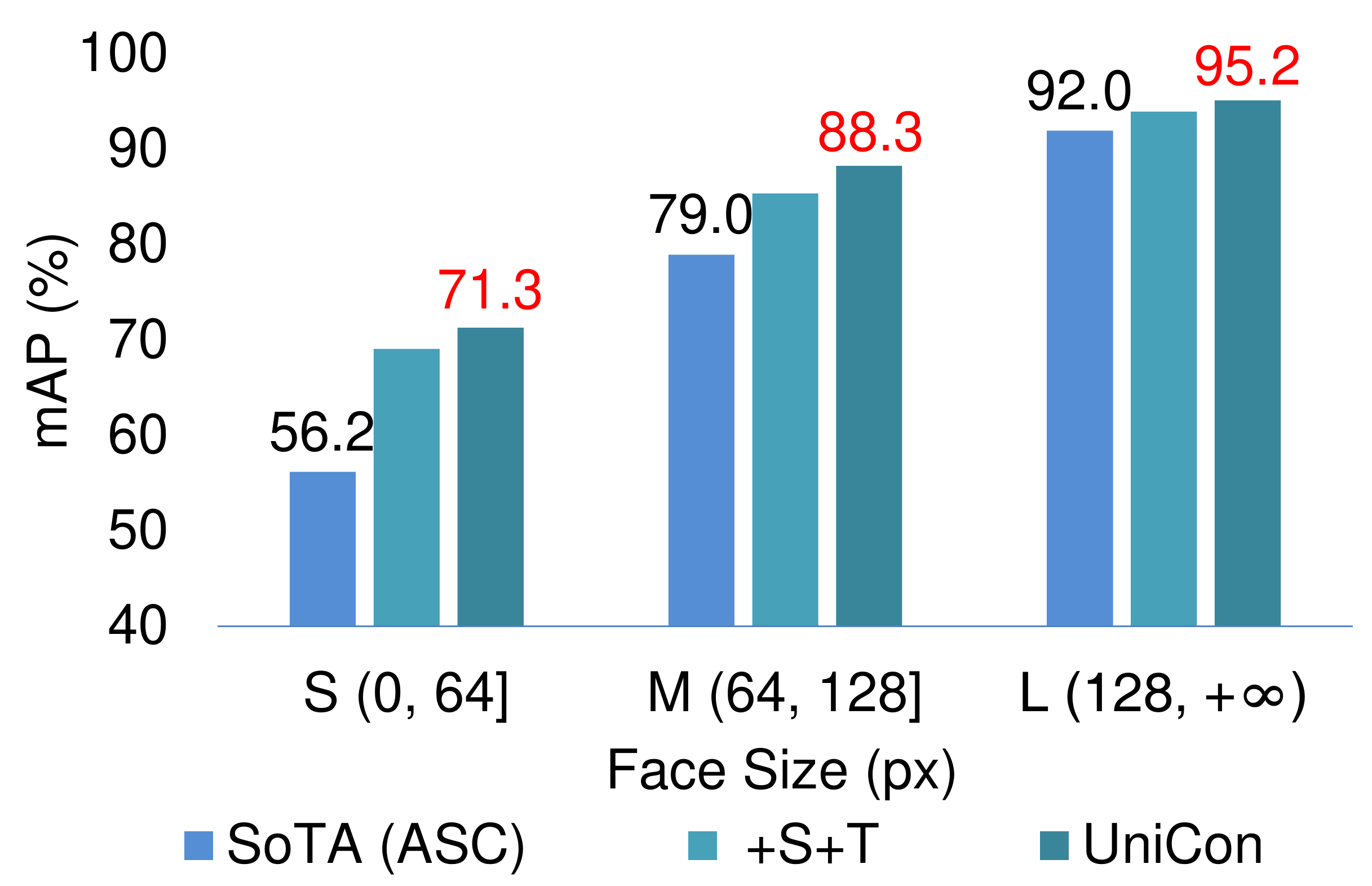}}
	\hfill
	\subfigure[mAP breakdown by number of on-screen faces.]{\includegraphics[width=0.48\columnwidth]{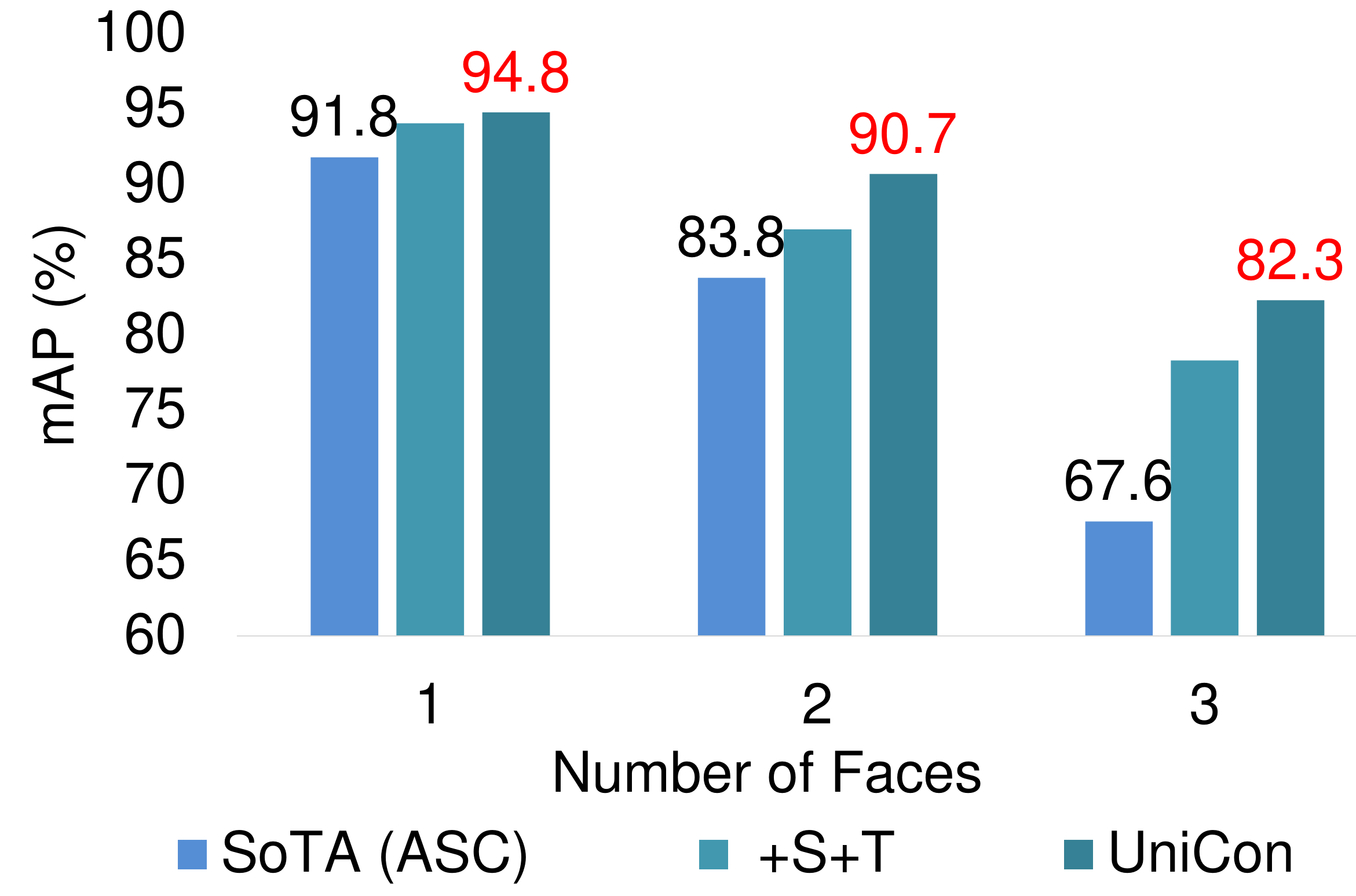}}
	\vskip-3.5ex
	\caption{\textit{Performance breakdown.} We evaluate the performance of our model on faces of different sizes and on frames with one, two, and three detected faces (see main text). The performance of the previous state-of-the-art~\cite{DBLP:conf/cvpr/AlcazarCMPLAG20} (ASC) and our best model are annotated for better comparison.}
	\label{fig:breakdown}
	\vskip-3.5ex
\end{figure}

\vspace{-1ex}
\paragraph{Multiple Candidates.} Fig.~\ref{fig:breakdown}b shows the model performance according to the number of detected faces in a frame. We consistently improve over the previous state-of-the-art for different numbers of detected faces, with the most significant gain being in the subset with $3$ faces ($+14.7$\% mAP). Moreover, we surpass $90\%$ on the two-faces subset for the first time. The improvements prove the effectiveness of spatial and relational context modeling, which equip the model with the ability to discern background actors from main actors, as well as an enhanced understanding of the relationships between the visible speakers. 
We refer interested readers to the supplemental video for more interesting qualitative results.
%We provide more qualitative results in Sec. \ref{sec:qualitative}.
%\input{figures/num_faces.tex}
\vspace{-1ex}
\paragraph{Out-of-sync Audio and Video.}\label{sec:out-of-sync}
Many real-world videos suffer from poor synchronization between audio and video, caused by transmission or re-encoding. To evaluate our model's performance on out-of-sync data, we assume perfect synchronization in the source videos and artificially shift the audio stream by up to $10$ frames to mimic out-of-sync videos. We then assess the performance of our model on manually de-synchronized videos. As shown in Fig.~\ref{fig:sync}, our model is fairly resilient to A-V sync errors. Remarkably, even on videos that are shifted by $10$ frames ($0.4$sec), our full \NetName{} model only loses $0.67$\% mAP, and still outperforms a simple Bi-GRU baseline by $1.64$\% absolute, which does not model spatial and relational context (\textbf{+T} in Table~\ref{table:ablations}), and receives \textit{synchronized} inputs. This shows that our model is robust to synchronization error.
\begin{figure}[!htb]
	\centering
	\vskip-2ex
	\includegraphics[width=0.62\columnwidth]{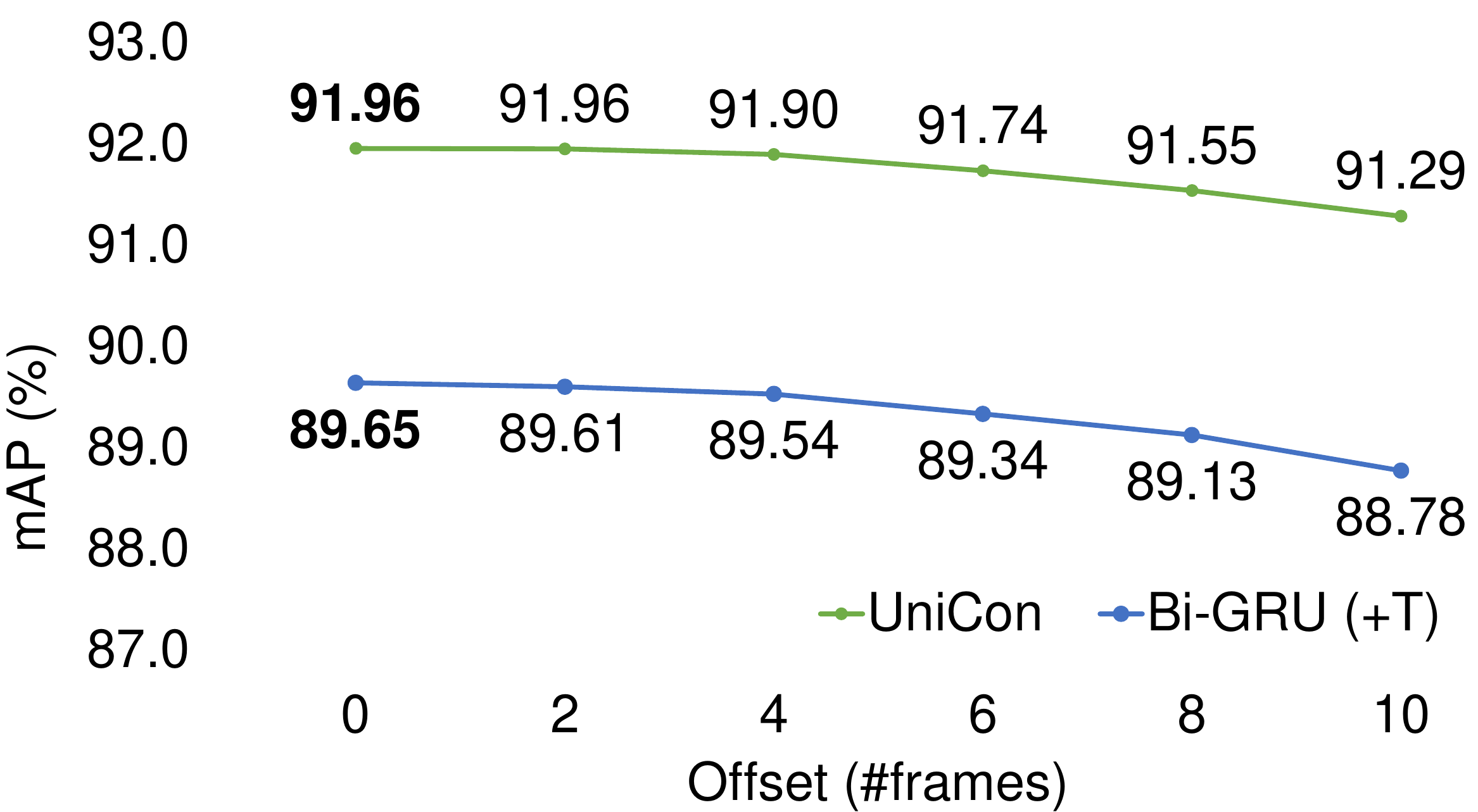}
	\vskip-2.5ex
	\caption{\textit{Results on out-of-sync videos.} Our model is fairly resilient to A-V synchronization error.}
	\label{fig:sync}
	\vskip-4ex
\end{figure}

\vspace{-1.5ex}

\section{Conclusion}
We have proposed a novel model named \NetName{} for active speaker detection. Key to our method is a unified modeling framework which efficiently aggregates different types of contextual evidence to make robust decisions, by considering the relationships between each candidate and others. We demonstrate via experiments on the large-scale AVA-ActiveSpeaker dataset and various other real-world datasets that our model successfully tackles challenging cases with multiple candidates and low-resolution faces, and outperforms state-of-the-art methods by a large margin.
\vspace{-1.5ex}

\section*{Acknowledgments}
This work was partially supported by the National Key R\&D Program of China (No. 2017YFA0700804) and the National Natural Science Foundation of China (No. 61876171).

\begin{Appendix}
	\section{AVA-ActiveSpeaker Dataset Statistics}
	Table~\ref{table:ava-statistics} provides some statistics about the AVA-ActiveSpeaker dataset. Note that the number of entities is a rough estimate of the upper bound, which is the number of merged face tracks that are obtained through the procedure described in Sec.~\ref{sec:setup} of the main paper.
	% \vspace{-2ex}
	\section{Cross-Dataset Evaluation Results}
	In this section, we provide detailed results for the cross-dataset evaluations in Sec.~\ref{sec:cross-dataset-eval} of the main paper. Results for Columbia, RealVAD and AVDIAR are presented in Table~\ref{table:columbia}, \ref{table:realvad}, and \ref{table:avdiar}, respectively. For an analysis of these results, please refer to Sec.~\ref{sec:cross-dataset-eval} of the main paper. Note that we find if the sampling rate of the videos are changed to 25fps (the value used during training), the results will be different. In these tables, we report the results at the original sampling rate. For fine-tuned results on RealVAD, we report the numbers from the best models on the validation set.
	
	\begin{table*}
	\centering
	\caption{\textit{AVA-ActiveSpeaker dataset statistics}. The number of entities and scenes are obtained after merging annotated face tracks as described in Sec.~4.1, and the last column indicates the total duration of all scenes in each partition. $^\ast$Test set labels are held out for a separate ActivityNet challenge.}
	\label{table:ava-statistics}
	\vskip-1ex
		\begin{tabular}{l|c|c|c|c|c}
			\toprule
			\textbf{Partition} & \textbf{\#Videos} & \textbf{\#Faces} & \textbf{\#Entities} & \textbf{\#Scenes} & \textbf{\#Hours} \\
			\midrule
			Train & $120$ & $2,676$k & $28,274$ & $15,499$ & $19.82$ \\
			Validation & $33$ & $768$k & $7549$ & $4310$ & $5.82$ \\
			Test$^\ast$ & $109$ & $2,054$k & $21,247$ & $13,529$ & $16.16$\\
			\bottomrule
		\end{tabular}
\end{table*}

	\begin{table*}
	\caption{\textit{Zero-shot results on the Columbia dataset}. We report $F$-1 scores (\%) for each speaker, and the overall average. The methods marked with an asterisk ($^*$) employ large-scale self-supervised pre-training.}
	\vskip-1ex
		\begin{tabular}{l|c|c|c|l|l|l}
			\toprule
			\multirow{2}{*}{\textbf{Method}} & \multicolumn{6}{c}{\textbf{Speaker}} \\ \cline{2-7} 
			& \textbf{Bell} & \textbf{Boll}    & \textbf{Lieb}    & \textbf{Long} & \textbf{Sick} & \textbf{Avg.} \\
			\midrule
			\multicolumn{7}{l}{\textit{Visual-only}} \\
			\midrule
			Cross-modal \cite{DBLP:conf/eccv/ChakravartyT16}
			& $82.9$ & $65.8$ & $73.6$ & $86.9$ & $81.8$ & $80.2$ \\
			Upper Body \cite{DBLP:conf/iccvw/ShahidBM19}
			& $87.3$ & $96.4$ & $92.2$ & $83.0$ & $87.2$ & $89.2$ \\
			S-VVAD \cite{Shahid_2021_WACV} & $92.4$ & $\bf 97.2$ & $92.3$ & $95.5$ & $92.5$ & $94.0$ \\
			\textbf{Ours}
			& $\bf 98.1$ & $87.9$ & $\bf 97.9$ & $\bf 99.3$ & $\bf 99.2$ & $\bf 96.5$ \\
			\midrule
			\multicolumn{7}{l}{\textit{Audio-visual}} \\
			\midrule
			SyncNet \cite{DBLP:conf/accv/ChungZ16a}$^\ast$
			& $\bf 93.7$ & $\bf 83.4$ & $86.8$ & $97.7$ & $86.1$ & $89.5$ \\
			LWTNet \cite{DBLP:conf/eccv/AfourasOCZ20}$^\ast$
			& $92.6$ & $82.4$ & $88.7$ & $\bf 94.4$ & $\bf 95.9$ & $90.8$ \\
			\textbf{Ours}
			& $93.6$ & $81.3$ & $\bf 93.8$ & $93.5$ & $92.1$ & $\bf 90.9$ \\
			\bottomrule
		\end{tabular}
	\label{table:columbia}
\end{table*}

	\begin{table*}
	\caption{\textit{Zero-shot and leave-one-out fine-tuning results on the RealVAD dataset}. We report $F$-1 scores (\%) for each panelist, the overall average, and the overall standard deviation.}
	\vskip-1ex
		\begin{tabular}{l|c|c|c|c|c|c|c|c|c|c|c}
			\toprule
			\multicolumn{1}{l|}{\multirow{2}{*}{\textbf{Method}}} & \multicolumn{11}{c}{\textbf{Speaker}} \\ \cline{2-12} 
			\multicolumn{1}{c|}{} & \textbf{P1}    & \textbf{P2}    & \textbf{P3}    & \textbf{P4}    & \textbf{P5}    & \textbf{P6}    & \textbf{P7}    & \textbf{P8}    & \textbf{P9}    & \textbf{Avg.}   & \textbf{Std.}
			\\ \midrule
			Beyan~\cite{9133504} (V)                 & $51.63$ & $53.49$ & $42.92$ & $51.70$  & $44.40$  & $50.48$ & $58.73$ & $67.94$ & $55.75$ & $53.04$ & $7.50$ \\
			Ours (V, zero-shot)           & $94.32$ & $73.98$ & $\bf 89.93$ & $76.73$ & $80.57$ & $\bf 93.61$ & $\bf 98.78$ & $83.51$ & $\bf 93.54$ & $87.22$ & $8.27$\\
			\textbf{Ours (V, fine-tuned)} & $\bf 96.47$ & $\bf 81.05$ & $86.93$ & $\bf 84.36$ & $\bf 89.94$ & $85.60$ & $94.92$ & $\bf 88.07$ & $90.90$ & $\bf 88.69$ & $\bf 4.66$\\
			Ours (AV, zero-shot)            & $86.72$ & $\bf 78.11$ & $70.48$ & $73.14$ & $68.93$ & $84.90$ & $93.01$ & $80.40$ & $86.99$ & $80.30$ & $7.82$ \\ 
			\textbf{Ours (AV, fine-tuned)} & $\bf 86.87$ & $76.45$ & $\bf 81.55$ & $\bf 86.98$ & $\bf 79.58$ & $\bf 88.86$ & $\bf 97.01$ & $\bf 84.50$ & $\bf 88.92$ & $\bf 85.64$ & $\bf 5.70$\\
			\bottomrule
		\end{tabular}
	\label{table:realvad}
\end{table*}

	\begin{table*}
	\caption{\textit{Zero-shot results on the AVDIAR dataset.} The average is computed over all test sequences. Diarization Error Rate (DER) (\%) is reported; the lower the DER is, the better.}
	\label{table:avdiar}
	\vskip-1ex
		\begin{tabular}{l|c|c|c|c}
			\toprule
			\multirow{2}{*}{\textbf{Subset}} & \multicolumn{4}{c}{\textbf{Methods}}                              \\ \cline{2-5} 
		   & \textbf{+S+T (V)} & \textbf{\NetName{} (V)}  & \textbf{+S+T (AV)} & \textbf{\NetName{} (AV)}\\ \midrule
			$1$ participant & $40.98$ & $\bf 39.94$ & $28.15$ & $\bf 26.87$ \\
			$2$ participants & $71.72$ & $\bf 67.52$ & $64.85$ & $\bf 61.03$ \\
			$3$ participants & $70.72$ & $\bf 63.02$ & $71.09$ & $\bf 60.97$ \\
			$4$ participants & $69.96$ & $\bf 67.47$ & $73.84$ & $\bf 67.45$ \\
			\midrule
			\textbf{Average} & $65.68$ & $\bf 61.08$ & $60.57$ & $\bf 55.16$
			\\
 			\bottomrule
		\end{tabular}
\end{table*}

\end{Appendix}

\bibliographystyle{ACM-Reference-Format}
\bibliography{refs}

\end{document}